\documentclass[11pt,a4paper]{article}
\usepackage[hyperref]{acl2021}
\usepackage{times}
\usepackage{latexsym}

\usepackage{microtype}

\aclfinalcopy % Uncomment this line for the final submission
\def\aclpaperid{483} %  Enter the acl Paper ID here

\usepackage[utf8]{inputenc}
\usepackage[T1]{fontenc}
\usepackage{amsfonts}

\usepackage{microtype}
\usepackage{graphicx}
\usepackage{subfigure}
\usepackage{booktabs}
\usepackage{enumitem}
\usepackage{amsmath}
\usepackage{url}
\usepackage{tabularx}

\usepackage{verbatim} 

\usepackage{multirow}
\usepackage{color}

\definecolor{ourlightblue}{HTML}{E0ECF7}
\definecolor{ourdarkblue}{HTML}{092E6B}
\definecolor{msgrblue}{HTML}{4889f4}
\definecolor{msgrgray}{HTML}{e1e1e7}

\definecolor{botc}{rgb}{0.458, 0.488, 0.978}

\definecolor{humanc}{rgb}{0.8, 0.8, 0.8}

\definecolor{light-gray}{gray}{0.90}
\definecolor{dark-gray}{gray}{0.30}

\definecolor{aszlam}{rgb}{1.0, 0.0, .6}

\title{Beyond Goldfish Memory\thanks{~~We use this term colloquially, see \citet{agranoff1965memory} for evidence of goldfish long-term memory.}: Long-Term Open-Domain Conversation}

\author{Jing Xu  \quad Arthur Szlam \quad Jason Weston\\
\\
 Facebook AI Research
}

\date{}

\begin{document}
\maketitle

\begin{abstract}
Despite recent improvements in open-domain dialogue models, state of the art models are trained and evaluated on short conversations with little context.
In contrast, the long-term conversation setting has hardly been studied. In this work we collect and release a human-human dataset consisting of multiple chat sessions whereby the speaking partners learn about each other's interests and discuss the things they have learnt from past sessions. 
We show how existing models trained on existing datasets perform poorly in this long-term conversation setting in both automatic and human evaluations, and we study long-context models that can perform much better.  In particular, we find  retrieval-augmented methods and methods with an ability to summarize and recall previous conversations outperform the standard encoder-decoder architectures currently considered state of the art.

\end{abstract}

\section{Introduction}

Improvements in the ability to train large neural language models, 
%Open-domain dialogue models are becoming more  with t
together with the availability of larger and higher quality dialogue datasets,  are spurring the development of increasingly convincing open-domain dialogue models.
Unfortunately, a major aspect missing from  the current state of the art is that human conversations can take place over long time frames, whereas  the currently used systems suffer in this setting.
%First, current 
Commonly used training and evaluation resources -- while large in terms of number of training examples -- include only short conversations, typically between 2-15 turns, consisting of a single conversational session. %and secondly, standard encoder-decoder architectures typically only encode 
Perhaps for that reason, 
%models have not been developed for this setting, e.g. 
the current state-of-the-art models such as Meena \cite{adiwardana2020meena} and BlenderBot \cite{roller2020recipes} employ Transformers with  token truncation lengths of only 128 tokens and are clearly incapable of incorporating long-term conversational context.  Consequently, it is unclear how well these models will perform on long or multi-session open-domain conversations.  In contrast, a successfully deployed bot will engage in many conversations over a length of time,  as capturing organic user interest will garner continual reengagement from returning users. Such long-term open-domain communication gives the opportunity for the conversation to develop and even improve with time as the model has more context and more understanding of that specific user's interests. 
In general, the standard encoder-decoder architectures currently used may not be sufficient in such a setup.

\begin{figure*}
    \centering
    \includegraphics[width=14cm]{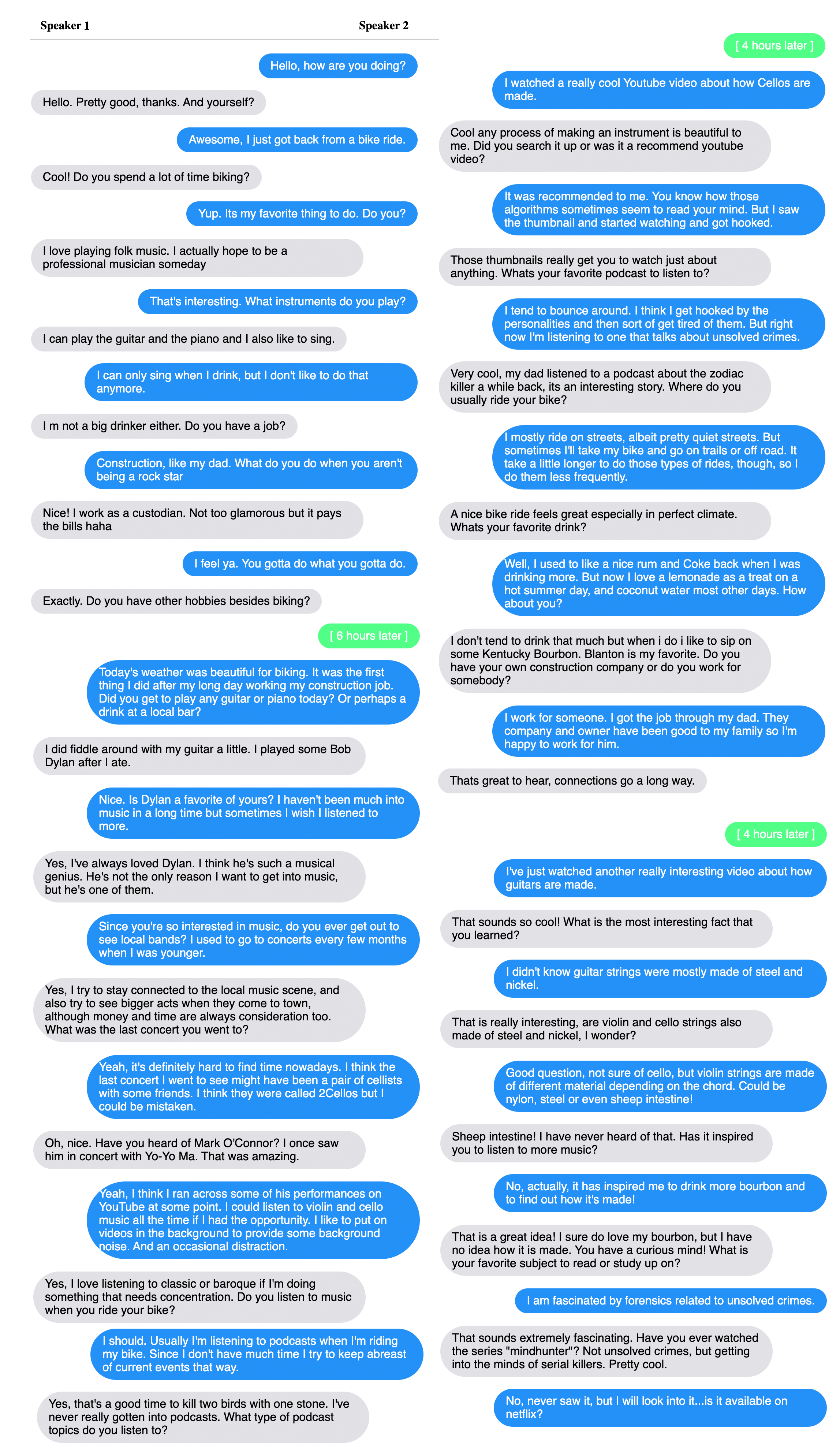}
    \caption{Example four session conversation from the newly collected Multi-Session Chat dataset. New sessions refer back to previous subjects, explore them in depth, or spark up conversation on new topics. }
    \label{fig:msc-example1}
\end{figure*}

In this work we study methods for long-term open-domain conversation. As to the best of our knowledge no public domain task exists to study such methods, we collect and release a new English dataset, entitled {\em Multi-Session Chat} (MSC). The dataset consists of human-human crowdworker chats over 5 sessions, with each session consisting of up to 14 utterances, where the conversationalists reengage after a number of hours or days and continue chatting. Previous sessions are annotated with summaries of important personal points that may be useful in further conversations. When reengaging, conversationalists often address existing knowledge about their partner to continue the conversation in a way that focuses and deepens the discussions on their known shared interests, or explores new ones given what they already know.
See  \autoref{fig:msc-example1} and  \autoref{fig:msc-example2} 
or  example conversations from our dataset.  

We study the performance of two long-context conversational architectures on this task: (i) retrieval-augmented generative models \cite{lewis2020retrieval,shuster2021retrieval}; and (ii) a proposed read-write memory-based model that summarizes and stores  conversation on the fly.
We show that both techniques outperform conventional encoder-decoder Transformers, and that training models on our new task give long-term conversational abilities that existing state-of-the-art models lack, as shown in both automatic metrics and human evaluations. We provide extensive experiments and ablations that study the reasons behind these improvements, and release models, data and code for researchers to evaluate further progress on this important problem\footnote{\url{http://parl.ai/projects/msc}}.

\begin{figure}
\end{figure}

% TODO update the numbers (break by session?)
\begin{table*}[h]
\center
\begin{small}
\begin{tabular}{lrrr|rrr|rrr}
 \toprule
            & \multicolumn{3}{c}{Train} & \multicolumn{3}{c}{Valid} & \multicolumn{3}{c}{Test}\\
 Data Type  & Epsiodes & Utts. & Summary & Epsiodes & Utts. & Summary & Epsiodes & Utts. & Summary \\
 \midrule
 Session 1  &  8939  &  131,438 & 59,894 & 1,000 & 7,801 & 7,768 & 1015  & 6,634  & 6,572 \\
 Session 2  &  4000  &  46,420  & 46,420 & 500 & 5,897 & 5,897 & 501 & 5,939 & 5,939\\
 Session 3  &  4000  & 47,259  & 26,976 & 500 & 5,890 & 5,890 & 501 & 5,924 & 5,924\\ 
 Session 4  &  1001  & 11,870  & - & 500 & 5,904  & 5,904 & 501 &  5,940 & 5,940 \\
 Session 5  &     -  & -  & - & 500 & 5,964  & - & 501 & 5,945  & - \\
 \midrule
 Total      &    -    & 236,987 & 133,290 & & 31,456  &25,459 & -  &30,382 &24,375 \\
\bottomrule  
\end{tabular}
\end{small}
\caption{
Data statistics of our {\sc Multi-Session Chat} dataset. Speakers converse across {\em sessions}, each of which is a short focused conversation, with subsequent sessions picking up the conversation again hours or days later.   We show the number of episodes, utterances (utts) and response summaries for each session. 
\label{tab:dataset}
}
\end{table*}

% TODO MSC, MSC3+
\begin{table*}[h]
\center
\begin{small}
\begin{tabular}{lrrrrrr}
\toprule
         & Num. &	Num.    	        &  Unique  &	Avg. Utt.	& Sessions & Utterances \\
Dataset	 & Episodes &	Utterances  & Tokens &	Length	& per Episode & per Episode \\
\toprule
Pushshift.io Reddit  &  - & 1.2B  & $\sim$1M & 25.4 &  1 & 3.2 \\
PersonaChat \tiny\cite{zhang2018personalizing} & 8,939 & 131,438 &  18,688 & 11.9 & 1 & 14.7\\
LIGHT  \tiny{\cite{urbanek2019learning}}	& 8,538 &	110,877		& 33,789 &	18.3 & 1 & 12.9 \\	
Wiz. of Wikipedia \tiny\cite{dinan2018wizard} & 18,430 & 166,787  &52,490  & 19.7 & 1 & 9.0 \\
Daily Dialog \tiny\cite{li2017dailydialog}            & 22,236 & 87,170 & 20,673 & 14.5 & 1 & 3.9 \\
Empathetic Dialog \tiny\cite{rashkin2019empathetic} &  24,850 & 64,636 &  19,458 &  15.3 & 1 & 2.6 \\
{\sc Multi-Session Chat} (1-3)  & 4,000	 & 161,440 & 37,366 & 21.4 & 3 & 40.4  \\
{\sc Multi-Session Chat} (1-4)  & 1,001	 & 53,332  & 23,387 & 23.0 & 4 & 53.3\\
\bottomrule  
\end{tabular}
\end{small}
\caption{
Comparison of the training data statistics of the {\sc Multi-Session Chat} (MSC) dataset  compared to several existing  datasets. We show MSC in two categories: episodes with 3 or 4 sessions, named (1-3) or (1-4).  
\label{tab:dataset_compare}
}
\end{table*}

\begin{table*}[h]
\center
\begin{small}
\begin{tabular}{lrrrrrrr}
\toprule
Pre-Train Model & Truncation &  Sessions 1-4 & Session 1  & Session 2 & Session 3 & Session 4 & Trunc\% (S4) \\
\toprule
\multicolumn{3}{l}{\em With no previous session context} \\
BST 2.7B & 128 &  9.23 & 8.76 & 9.45 & 9.31 & 9.40 & 51\% \\
BST 2.7B & 512 &  9.06 & 8.18 & 9.42 & 9.26 & 9.36 & 0\% \\
BST 2.7B & 1024 & 9.08 & 8.20 & 9.46 & 9.29 & 9.37 & 0\% \\
%BART 400M & 1024 &  \\
\midrule
\multicolumn{3}{l}{\em With previous session dialogue  context} \\
%Reddit 2.7B & 128 & 9.20 & 9.00 & 9.04 & 9.36 &  9.40  \\
%Reddit 2.7B & 512 & 8.79 & 8.08 & 8.83 & 9.08 & 9.20\\
%Reddit 2.7B & 1024 & 8.91& 8.29 & 8.79 & 9.24 &  9.31\\
BST 2.7B & 128 &  9.16 & 8.75 & 9.32 & 9.22 & 9.32 & 100\%\\
BST 2.7B & 512 & 8.87 & 8.15  & 9.14 & 9.04 & 9.17 & 100\%\\
BST 2.7B & 1024 & 8.89& 8.17  & 9.18 & 9.05 & 9.16 & 80\%\\
%BART 400M & 1024 &  \\
\midrule
\multicolumn{3}{l}{\em With previous session summary context} \\
BST 2.7B & 128 & 9.09 & 8.77 & 9.24 & 9.12 & 9.24 & 100\% \\
BST 2.7B & 512 & 8.79 & 8.17 & 8.69 & 9.15 & 9.22 & 36\% \\
BST 2.7B & 1024 & 8.80 & 8.18 & 9.05 & 8.91 & 9.04& 0\% \\
%BART 400M & 1024 &  \\
\bottomrule  
\end{tabular}
\end{small}
\caption{
{\bf Comparison of different context truncation lengths and context types} when training on {\sc Multi-Session Chat}. We show validation perplexity for various models across different sessions, and percent of tokens truncated for session 4 (last column). 
\label{tab:context_length_and_type}
}
\end{table*}

\begin{table*}[h]
\center
\begin{small}
\begin{tabular}{lrrrr|rrrr}
\toprule
                & \multicolumn{4}{c}{Session } &  \multicolumn{4}{c}{Session Openings}\\
 Model Context     &    2 &   3 & 4 & 5 &  2 &   3 & 4 & 5 \\
\toprule
No Session History                   &   9.46 & 9.29 & 9.37 & 9.30  &  9.96 & 10.99& 10.69 &  10.46 \\
Dialogue History                     &   9.18 & 9.05 & 9.16 & 9.08 &  7.55 & 8.48 & 8.27 & 7.94\\ 
Gold summary                         &   9.04 & 8.90 & 9.02 & 8.96 &  6.98 & 7.96 & 7.94 & 7.77\\
Gold summary (without time features) &   9.05 & 8.91  & 9.04 & 8.95 &  6.97 & 7.95 & 7.97 & 7.74 \\
Gold summary (partner's only)        &   9.14 & 8.99 & 9.11 & 9.03 & 7.66   & 8.49 & 8.49 & 8.07\\
Gold summary (self only)             & 9.29    &  9.10 &  9.18 &  9.13 &  8.40 & 8.94 & 8.52 & 8.39\\
Predicted Summary                 &   9.11 & 8.98 & 9.07 & 9.00  & 7.44 & 8.43 & 8.20 & 7.81\\
\bottomrule  
\end{tabular}
\end{small}
\caption{
{\bf Summaries vs. Dialogue Context Performance} when training on {\sc Multi-Session Chat}, reporting validation perplexity, using a BST 2.7B-1024 pre-trained model with MSC fine-tuning.  Note that the last row in this Table corresponds to the SumMem-MSC 2.7B (truncate 1024) row in Table \ref{tab:main_results_valid} in the Appendix.
\label{tab:summary_vs_dialog}
}
\end{table*}

\begin{table*}[h]
\center
\begin{small}
\begin{tabular}{lrrrr|rrrr|r}
\toprule
                & \multicolumn{4}{c}{Session } &  \multicolumn{4}{c}{Session Openings}\\
 Model Context     &    2 &   3 & 4 & 5 &  2 &   3 & 4 & 5  & Sparsity\\
\toprule
Gold summary                         &   9.04 & 8.90 & 9.02 & 8.96 &  6.98 & 7.96 & 7.94 & 7.77 & 42.0\% \\
\midrule
Predicted Summary  (sampling 5\%)    &   9.11 & 8.98 & 9.07 & 9.00  & 7.44 & 8.43 & 8.20 & 7.81  &  29.1\%\\
Predicted Summary  (sampling 25\%)   &   9.11 & 8.97 & 9.07 & 9.01  &  7.46  & 8.53 & 8.22 & 7.94 & 41.4\% \\
Predicted Summary  (sampling 50\%)   &   9.14 & 8.99 & 9.08 & 9.02  &  7.57 & 8.62 & 8.37 & 8.11 & 50.7\%\\
Predicted Summary  (sampling 100\%)     &   9.14   & 8.99 & 9.10   &  9.03  & 7.68 & 8.69 &  8.56 &8.25 & 61.8\%\\
\bottomrule  
\end{tabular}
\end{small}
\caption{
{\bf Predicted Summaries when subsampling the no-summary class}  on {\sc Multi-Session Chat}, reporting validation perplexity, using a BST 2.7B-1024 pre-trained model with MSC fine-tuning. The last column shows the sparsity of the summarizations (how often a summary line is generated), which can be controlled by subsampling the no-summary class at training time. Subsampling gives better results and closer sparsity levels to the original human annotated data.
\label{tab:summary_sampling}
}
\end{table*}

\begin{table}[h]
\center
\begin{small}
\begin{tabular}{lrrrrrr}
\toprule
                & \multicolumn{5}{c}{Session } \\
 Training Data  &  1 &  2 &   3 & 4 & All\\
\toprule
%Session 1         & 8.34 & 10.6 &11.12 & 11.2   & 10.3 \\
%Sessions 1+2      & 8.25 & 8.88 & 9.36 & 9.41    & 8.97\\
%Sessions 1+2+3    & 8.25 & 8.70 & 9.21 & 9.25    & 8.85\\
%Sessions 1+2+3+4  & 8.21 & 8.67 & 9.16 & 9.23  & 8.82 \\
Session 1         & 8.24 & 11.4 & 11.2 &  11.3   & 10.5 \\
Sessions 1+2      & 8.21 & 9.21 & 9.09 & 9.24    & 8.94 \\
Sessions 1+2+3    & 8.16 & 9.05 & 8.93 & 9.06    & 8.80 \\
Sessions 1+2+3+4  & 8.16 & 9.02 & 8.89 & 9.02  & 8.77 \\
\bottomrule  
\end{tabular}
\end{small}
\caption{
{\bf Varying the Number of Training Sessions} when training on {\sc Multi-Session Chat}, reporting validation perplexity, using a BST 2.7B-1024 pre-trained model with MSC  using gold summaries. 
\label{tab:vary_train_sessions}
}
\end{table}

\begin{table*}[h]
\center
\begin{small}
\begin{tabular}{lrrrrrrr}
\toprule
%Model & Session 1 & Session 2 & Session 3 & Session 4 & Session 5 & Session Openings(2-4) \\
Model & Session 1 & Session 2 & Session 3 & Session 4 & Session 5 & Session Openings \\
\toprule
%BST 2.7B \cite{roller2020recipes}  & 8.84 & 10.56& 10.44&10.51 & 10.44 & 13.04 \\ {valid} 
BST 2.7B \cite{roller2020recipes}  & 8.97 & 9.98 & 10.26 & 10.40 & 10.50 &  12.92 \\
%{\sc MSC} 2.7B (truncate 128)      & 8.75 & 9.32 & 9.22 & 9.32 & 9.23 & 8.95  \\ {valid} 
{\sc MSC} 2.7B (truncate 128)      &8.87 & 8.89 & 9.10 & 9.21 & 9.27 &  8.95 \\
%{\sc MSC} 2.7B (truncate 1024)     & 8.17 & 9.18 & 9.05 & 9.16 & 9.08 & 8.06 \\ {valid} 
{\sc MSC} 2.7B (truncate 1024)     & 8.25 & 8.76  & 8.93 & 9.07 & 9.16 & 8.09 \\
% BST 2.7B + {\sc MSC} fine-tune (128)   &  \\
%BST 2.7B + {\sc MSC} fine-tune (1024)     &  \\
% MSC 2.7B (Reddit + MSC fine-tune)                          & \\
\midrule
% (1 document = 1 utterance) RAG-{\sc MSC} 2.7B  &    8.36 &  9.37 & 9.21 & 9.3   & 9.26 & ?   \\
%{\sc MSC} 2.7B (RAG)  &  8.14 &  9.16 & 9.06 & 9.18   & 9.10 & 8.04 \\ % (1 document = 1 session dialogue history) {valid} 
{\sc MSC} 2.7B (RAG)  &  8.22 &  8.78 & 8.97 & 9.11   & 9.17 & 8.10 \\ % (1 document = 1 session dialogue history)
%{\sc MSC} 2.7B (FiD) &  8.16 &  9.14 & 9.02 & 9.10   & 9.04 &  7.97  \\ % (1 document = 1 session dialogue history) {valid} 
{\sc MSC} 2.7B (FiD) &  8.22 &  8.75 & 8.92 & 9.05   &  9.11 & 8.06  \\ % (1 document = 1 session dialogue history)
%{\sc MSC} 2.7B (FiD-RAG) &  8.16 &  9.13 & 9.02 & 9.10   & 9.04 & 7.96   \\ % (1 document = 1 session dialogue history) {valid} 
{\sc MSC} 2.7B (FiD-RAG) &  8.23 &  8.75 & 8.93 & 9.04   & 9.11  & 8.03   \\ % (1 document = 1 session dialogue history)
\midrule
%SumMem-{\sc MSC} 2.7B (truncate 1024) &  8.18 &  9.11 & 8.98 & 9.07 & 9.00 & 7.97 \\ {valid} 
SumMem-{\sc MSC} 2.7B (truncate 1024) &  8.25 &  8.71 & 8.89 & 9.01 & 9.09 & 8.04 \\
%SumMem-{\sc MSC} 2.7B (RAG)     &   8.16 &	9.19 &	9.07& 	9.17 & 9.09 & 7.95 \\ {valid} 
SumMem-{\sc MSC} 2.7B (RAG)     &   8.24 &	8.81 &	9.00& 	9.10 & 9.17 & 8.05 \\
%SumMem-{\sc MSC} 2.7B (FiD)     & 8.16	& 9.09 	& 8.97	& 9.07 & 8.99 & 7.82   \\ {valid} 
SumMem-{\sc MSC} 2.7B (FiD)     & 8.20	& 8.71 	& 8.89	& 9.00 & 9.07 & 7.91   \\
%SumMem-{\sc MSC} 2.7B (FiD-RAG) & 8.16 & 9.08 & 8.96 & 9.07 & 8.99 & 7.78   \\ {valid} 
SumMem-{\sc MSC} 2.7B (FiD-RAG) & 8.22 & 8.70 & 8.89 & 9.00 & 9.07 &  7.87   \\
\bottomrule  
\end{tabular}
\end{small}
\caption{
{\bf Test perplexity across sessions} for our retrieval- and memory-augmented models  (bottom two blocks) compared to several encoder-decoder baselines (top three rows).
\label{tab:main_results_test}
}
\end{table*}

\section{Related Work}

A relatively large and growing number of either natural or crowdsourced datasets have been collected and used in open-domain dialogue research. These datasets focus on the vast array of different skills required by a dialogue agent, but conversations lengths are typically short. Recent state-of-the-art open-domain dialogue agents have utilized  Daily Dialogue \cite{li2017dailydialog}, PersonaChat \cite{zhang2018personalizing}, Empathetic Dialogues \cite{rashkin2019empathetic}, Wizard of Wikipedia \cite{dinan2018wizard} and Pushshift.io Reddit \cite{baumgartner2020pushshift}; see \citet{huang2020challenges} for a review of other datasets.
The number of conversational turns in these datasets is in the range of 2-15 turns, we provide statistics of some of these datasets in \autoref{tab:dataset_compare}. Crowdsourcing long conversations is difficult due to both the expense and the difficulty of employing crowdworkers for long lengths of time due to so called Human Intelligence Tasks (HITs) being typically of a short duration -- only ``a few minutes'' \cite{paolacci2010running}.
While organic long conversations regularly transpire on the internet, e.g. on messaging platforms, these are  proprietary, and privacy concerns make public release implausible.

Several existing datasets explore the use of personal knowledge used as context to dialogue, which can be seen as a short, simple memory provided to the bot.
%Personal knowledge used as context to dialogue has been explored in a number of works, and can be seen as a short, simple memory provided to the bot.
In \citet{mazare2018trainingmillions} such personas were extracted from Reddit and used to train agents. In \citet{zhang2018personalizing} personas were first crowdsourced, and speakers were asked to play those roles. Other works have considered encoding personas into vector-based weights \cite{li2016persona}.

In this work, we explore summarizing the long-term conversations that occur in order to store useful information about them.
Summarization is a rich field where the vast majority of work focuses on summarizing documents \cite{kaikhah2004automatic,kryscinski2019neural,cheng2016neural}, for example summarizing in order to predict other relevant information \cite{west2019bottlesum},
and there is some work on dialogue as well \cite{goo2018abstractive,gliwa2019samsum,pan2018dial2desc}. 

Standard Transformers have a fixed context length which due to the all-vs-all self-attention mechanism becomes inefficient when it is too large. Consequently, many existing pre-trained models have short token truncation lengths, e.g. 128 tokens, as in BlenderBot \cite{roller2020recipes} and Meena \cite{adiwardana2020meena}, or 1024 tokens, as in BART \cite{lewis-etal-2020-bart}. A number of approaches have been proposed to ameliorate this issue.
Long-context Transformers consider ways to speed up the self-attention mechanism  \cite{child2019generating,kitaev2019reformer,beltagy2020longformer} and retrieval-augmented methods consider ways to select the pertinent parts of the context to keep in the considered set of tokens
\cite{dinan2018wizard,lewis2020retrieval,shuster2021retrieval} which can be linked to earlier methods in memory networks \cite{weston2014memory} and neural QA \cite{chen2017reading}.
We consider some of these approaches in this work.

\section{Multi-Session Chat}

To conduct research on long-term conversations, we require data to both train on and to evaluate models. 
Rather than attempting to collect a set of dialogues with one single long conversation per speaker pair,  we consider the natural case where two speakers chat online in a series of sessions as is for example common on messaging platforms. Each individual chat session is not especially long before it is ``paused''. Then,  after a certain amount of (simulated) time has transpired, typically hours or days, the speakers resume chatting, either continuing to talk about the previous subject, bringing up some other subject from their past shared history, or sparking up conversation on a new topic. We consider this multi-session long conversation setup, and name our dataset 
{\em Multi-Session Chat} (MSC).

\paragraph{Crowdworker Data Collection} To build a publicly available dataset, as in other datasets, we employ crowdworkers to engage in open-ended chat. The crowdworkers act in the roles of speakers engaged in open-ended multi-session chat spanning hours, days or weeks (however, note that those lengths of time do not actually transpire in reality before the next chat session begins).

\paragraph{Personas} Crowdworkers are asked to play a role, rather than speaking about their own personality, which helps mitigate privacy converns, and ensures diversity even if the same crowdworker conducts multiple conversations. In addition to the crowdworkers being specifically told to play the role, they are also told not to discuss aspects of their real profiles or indeed any personally identifiable information. The role is provided as a series of sentences describing characteristics, events and opinions of the character they are playing. 
We use the 1155 personas crowdsourced from \citet{zhang2018personalizing}, validation and test use separate  personas from the ones used in the training set.
%For training take a random subset of 1155 personas crowdsourced from \citet{zhang2018personalizing}, validation and test use separate  personas from training. %Because the personas are not the real profiles of the crowdworkers, the dataset should not contain personal identifiable information (and they are told specifically to play the given roles and not to use any during any followup sessions).

\paragraph{Session 1}
For the first chat session we use the existing {\sc PersonaChat} dataset \cite{zhang2018personalizing}, which already involves short conversations where two speakers get to know each other for the first time. We note that these conversations rarely go beyond the superficial stage because speakers simply do not have enough turns to discuss any topic deeply.

\paragraph{Sessions 2, 3, 4, \dots}

To model subsequent sessions, we first select a random amount of time that has elapsed since the previous session, chosen to be either 1-7 hours or 1-7 days, as ideally speakers would reengage within that timeframe.
We ask the crowdworkers to play the same roles that were played in the previous session, acting as if that amount of time has transpired. We note these crowdworkers may not be the same ones that played those characters in previous sessions, but will be playing the same roles: this makes the task tractable in a crowdworking frameworking where jobs are typically short, and matching pairs over a long duration would be infeasible. We instruct the workers to 
``chitchat with another worker for 6 turns, as if you were {\em catching up} since last time you two spoke.'' and that ``When you expand the topic, make sure it makes sense with the personal details {\em already} mentioned.'', i.e. emphasizing that not only must they play their role, but also pay attention to previous interactions with the other speaker.

\paragraph{Session Lengths}  We collect two lengths of training conversation: 4000 episodes with 3 sessions, and 1001 episodes with 4 sessions. For the validation and test data, the sessions extend up to 5 sessions, giving us a way to measure long-context session performance that extends beyond the training set 
distribution.

\paragraph{Conversation Summaries (Extended Personas)}

We give crowdworkers access to all previous dialogues between the two conversational roles (for the role they are playing, and their partner's role). However, as the conversation gets longer, this becomes infeasible to read and digest within a limited amount of time. Instead, between each session, including after session 1, we run a separate crowdworker task in which conversations are summarized into important points. We then show these summaries as the primary reference for subsequent session dialogues, which are much shorter than the full dialogues themselves. As  these summaries were collected in order to store the important points pertinent to either one or the other speaker, they can also be seen to function as extensions of the original given personas. As the two speakers continue to converse they create more depth to those characters.

\paragraph{Dataset Examples} We show two dataset examples in \autoref{fig:msc-example1} and \autoref{fig:msc-example2} which consist of four sessions. We also show example summary annotations
in \autoref{fig:summary-exs}.

\paragraph{Dataset Statistics}
Statistics of the multi-session chat dataset are given in \autoref{tab:dataset} and a comparison with  other standard open-domain dialogue datasets is also given in \autoref{tab:dataset_compare}.
%We can see that the data is large in terms of raw numbers of  utterances compared to other open-domain dialogue datasets commonly used, see \autoref{tab:dataset_compare}. Most importantly 
We can see that the number of training utterances per episode is larger than other datasets (last column of Table \ref{tab:dataset_compare}). Our multi-session training chats that last 4 sessions have an average of $\sim$53 utterances in a given full conversation (over all sessions), while our validation and test chats over 5 sessions have an average of 
$\sim$66 utterances. In contrast,
other standard datasets are in the range of 2.6-14.7 utterances on average.
This brings new challenges in dialogue modeling due to the large context size, e.g. an average of 1614 tokens as tokenized by the BlenderBot BPE dictionary \cite{roller2020recipes}, where the Transformer used in that work has a truncation length of 128. Our dataset can be used both to  train  long-context conversational models, and also allows to evaluate such models, which was previously not easily possible.

\section{Modeling Multi-Session Chat}

\subsection{Transformer Encoder-Decoders}

The most straight-forward approach for modeling dialogue using our new task is simply to use a large language model as is standard in open-domain dialogue, i.e. an encoder-decoder Transformer as in the Meena \cite{adiwardana2020meena} and BlenderBot \cite{roller2020recipes} systems. We consider using the BST 2.7B parameter model from BlenderBot 
%and the 400M parameter BART model \cite{lewis-etal-2020-bart} 
as an initial pre-trained model, which we then fine-tune on the Multi-Session Chat task.

\paragraph{Encoder Truncation}
As BST 2.7B has a truncation of 128 tokens in the encoder, we consider extending this to a larger input. To do this, we extend its learnable positional encodings from 128 to 256, 512 or 1024 tokens, and then train these extra parameters at the same time as we fine-tune the whole network on the downstream task. 
We add new positional embeddings to be trained such that the existing ones (the first 128) do not change from before.
We then evaluate the impact of these modifications in order to select the best model. 

\subsection{Retrieval-Augmentation} \label{sec:retrieval-aug}

A popular technique for employing a Transformer encoder with a large context, only some of which is relevant, is to use retrieval augmentation. In this method, a retrieval system is used to find and select part of the context to be included in the final encoding which is attended to by the decoder. 

\paragraph{RAG}
The RAG (Retrieval-Augmented Generation) approach \cite{lewis2020retrieval} utilizes a neural-retriever-in-the-loop which is itself a Transformer
to do this. 
Documents or passages to be retrieved are stored in an approximate nearest-neighbor FAISS index \cite{johnson2019billion}, 
and  a DPR (Dense Passage Retrieval) \cite{karpukhin2020dense} Transformer bi-encoder model is used to  score document-context pairs in order to rank them based on their match, where the base DPR model is pre-trained on QA data pairs.
The DPR model is thus  used to both retrieve from the FAISS index, and then score the top $N$ candidates.  The entire system is trained end-to-end so that retrieval is optimized to help improve generation. This setup was shown to work for dialogue in particular in \citet{shuster2021retrieval}. 

\paragraph{FiD and FiD-RAG}
We also consider the Fusion-in-Decoder (FiD) \cite{izacard2020leveraging}, another method that has been shown to perform well. 
In this approach, the pre-trained retriever is used directly: 
each of the top $N$ documents returned is prepended to the context and encoded separately by the encoder, and finally all the results are concatenated. The decoder then attends to these encodings to produce a final response. We consider the pre-trained retriever to either be standard pre-trained DPR, or the RAG-trained retriever, called FiD-RAG \cite{shuster2021retrieval}.

\paragraph{Retriever and Documents}
In this work the set of passages in the memory is not large enough to require a FAISS index, but it is large enough that retrieval may be useful. We thus store for every item in the memory the vector encoding by the DPR model (whereas in the FAISS approach this dense vector is approximated instead). Then given a dialogue context, we score each memory 
using the bi-encoder, and use the top $N$ for generation. In our case, the memories consists of dialog utterances from the history of the conversation. We consider the chunk (passage) size as a hyperparameter and try either encoding  utterances as separate documents, or else whole sessions (or session summaries) as documents. The latter (whole sesions) worked better, and we report those in the final results. For $N$ we try values 3, 5 and 6, and also choose the best for each method according to the validation set.

\subsection{Summarization Memory-Augmentation}

The retrieval-augmentation model described in the previous section retrieves from the set of past dialogues. 
Simply storing historical context in the memory in its raw form is a simple approach that is often used elsewhere in the literature, e.g. in question answering or knowledge-grounded dialogue. However, those approaches have two potential drawbacks: (i) there is a lot of context to store, and hence retrieve from; (ii) no processing has been done on that content, so the reading, retrieving and combining to finally generate leaves a lot of work for the model to do.
We therefore propose instead a memory augmentation that first summarizes the {\em pertinent} knowledge and only stores that instead in an attempt to solve both problems.

The procedure involves two main components:
\begin{enumerate}
    \item An encoder-decoder abstractive summarizer that takes as input the dialogue history with the goal of summarizing any new pertinent information contained in the last dialogue turn. This includes the case of deciding that no new information should be stored in the memory. When found, the summarized knowledge is added to the long-term memory.
    \item A memory-augmented generator that takes the dialogue context and access to the long-term memory, and then generates the next response.
\end{enumerate}

For (1) we can use the human annotated data from our newly collected MSC task to know what summaries to generate. We thus train a supervised encoder-decoder model to produce summaries.

For (2) we can use the same systems as presented in \autoref{sec:retrieval-aug} to both retrieve from the summarization memories, and to finally generate an appropriate response.

\begin{table*}[bht!]
    \centering
    \small
    \begin{tabular}{lrrrr|rr}
            	          & Reference   & Reference     & New   &   Engaging  & Final   & \# Annotated \\
        Model	          & own topic  & other's topic   & topic &  Response &  Rating   & Responses\\
\hline
BST 2.7B \cite{roller2020recipes}  & 19.9\% & 14.5\%  & 69.0\% & 53.0\% & 3.14 & 668  \\
\hline
MSC 2.7B (truncate 128)            & 15.8\% & 21.8\%  & 75.8\% & 56.5\% & 3.29 & 673   \\
MSC 2.7B (truncate 1024)           & 15.0\% & 22.5\%  & 74.4\% & 54.2\% & 3.47 & 653 \\
\midrule
SumMem-{\sc MSC} 2.7B (RAG)        & 19.6\% & 33.8\%  & 72.7\% & {\bf 62.1\%} & {\bf 3.65} & 668 \\
SumMem-{\sc MSC} 2.7B (FiD)        & 22.1\% & 30.7\%  & 76.4\% & {\bf 58.9\%} & {\bf 3.62} & 662 \\
SumMem-{\sc MSC} 2.7B (FiD-RAG)    & 24.2\% & 26.4\%  & 78.3\% & {\bf 59.3\%} & {\bf 3.68} & 649 \\

\end{tabular}
\caption{{\bf Human Evaluation Results.} Performance of various models measured during conversations with crowdworkers.
Engaging response and final rating numbers in bold are statistically significant compared to BST 2.7B ($p$-value $< 0.05$) using a $t$-test.
}  
\label{tab:human_eval}
\end{table*}

\begin{figure}
    \centering
    \includegraphics[width=6.5cm]{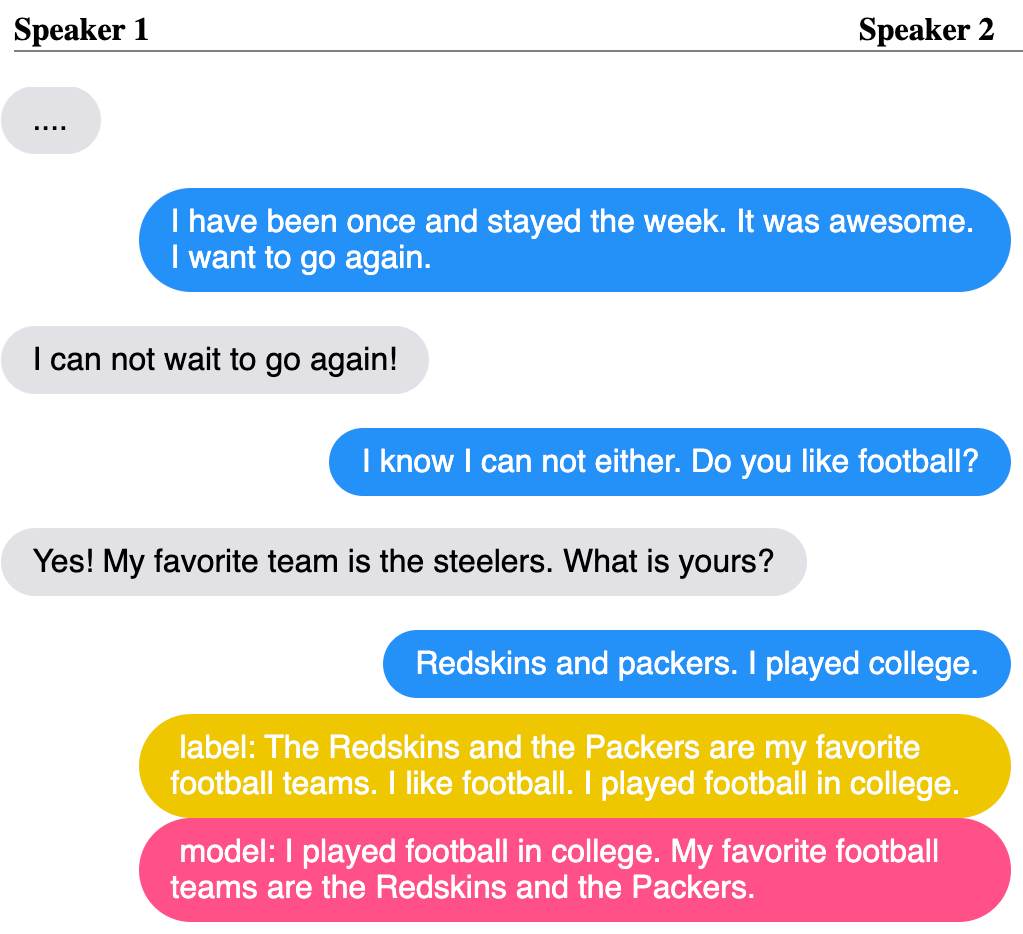}
    \includegraphics[width=6.5cm]{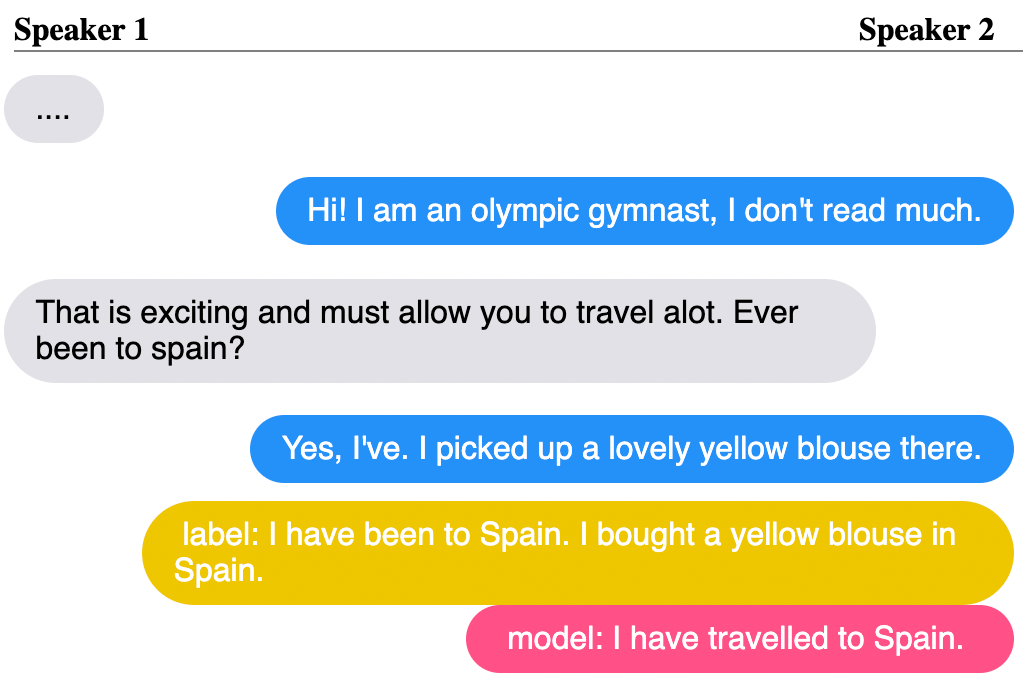}
    \includegraphics[width=6.5cm]{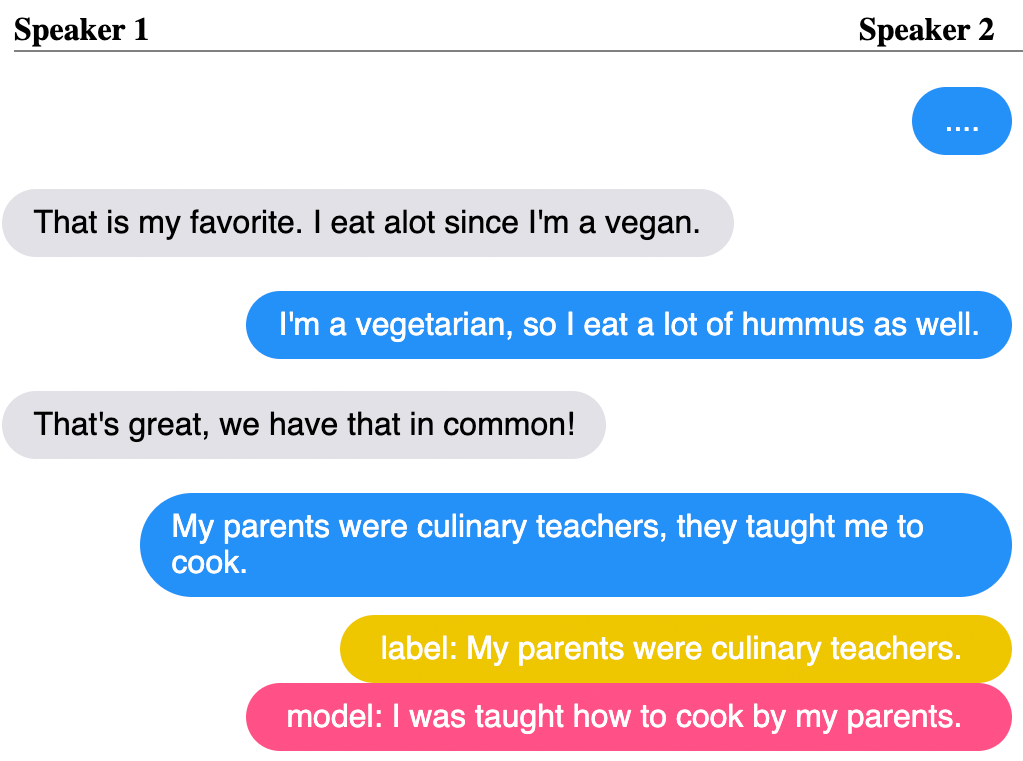}
    \includegraphics[width=6.5cm]{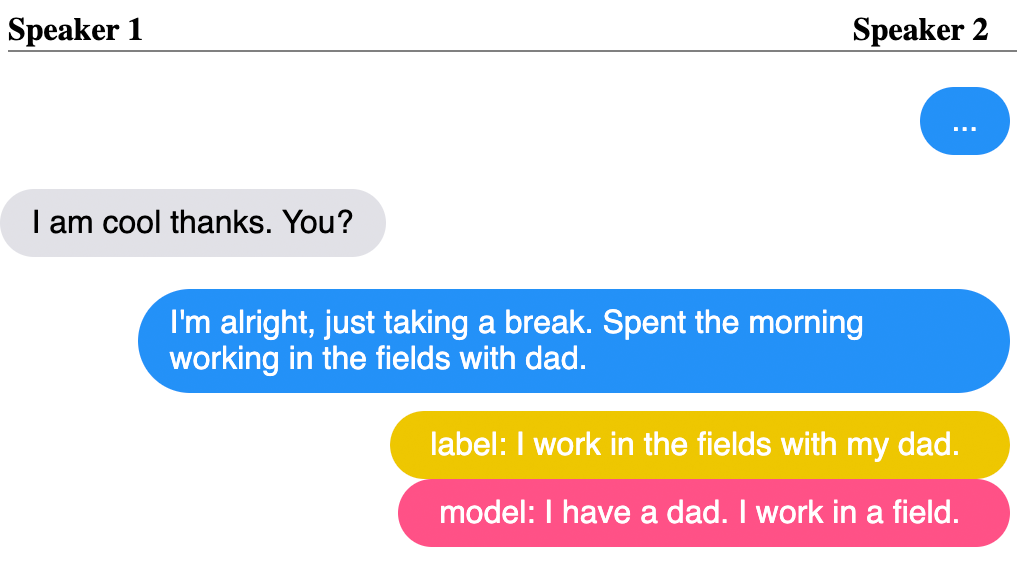}
    \includegraphics[width=6.5cm]{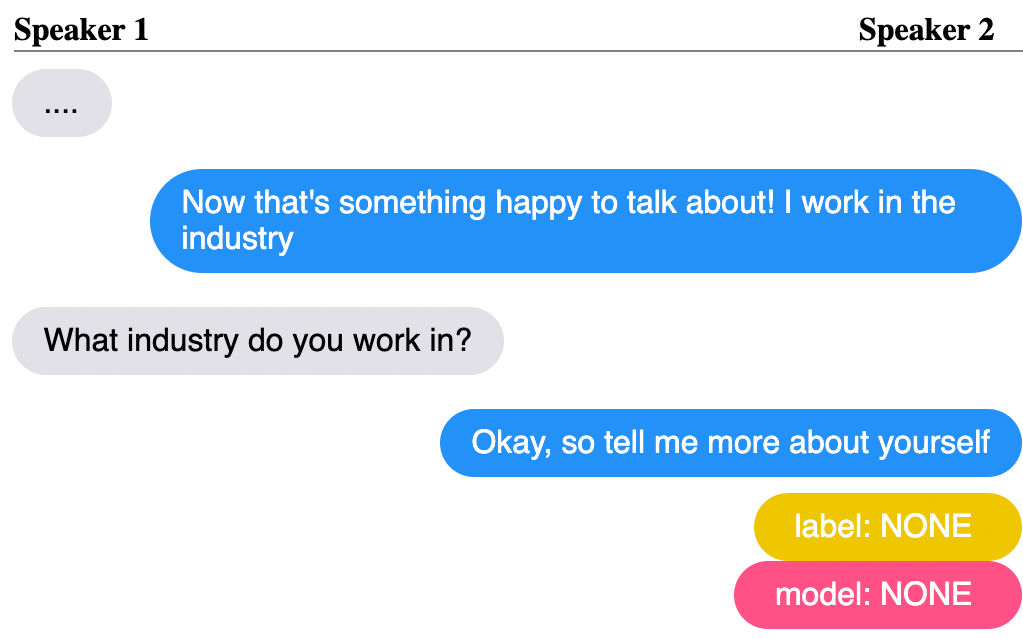}
    \caption{Example summary annotations and predictions on the validation set. We show the gold human annotation (label) and our model prediction (model).}
    \label{fig:summary-exs}
\end{figure}
\begin{figure}
\end{figure}

\begin{figure}[bht!]
    \centering
    \includegraphics[width=7cm]{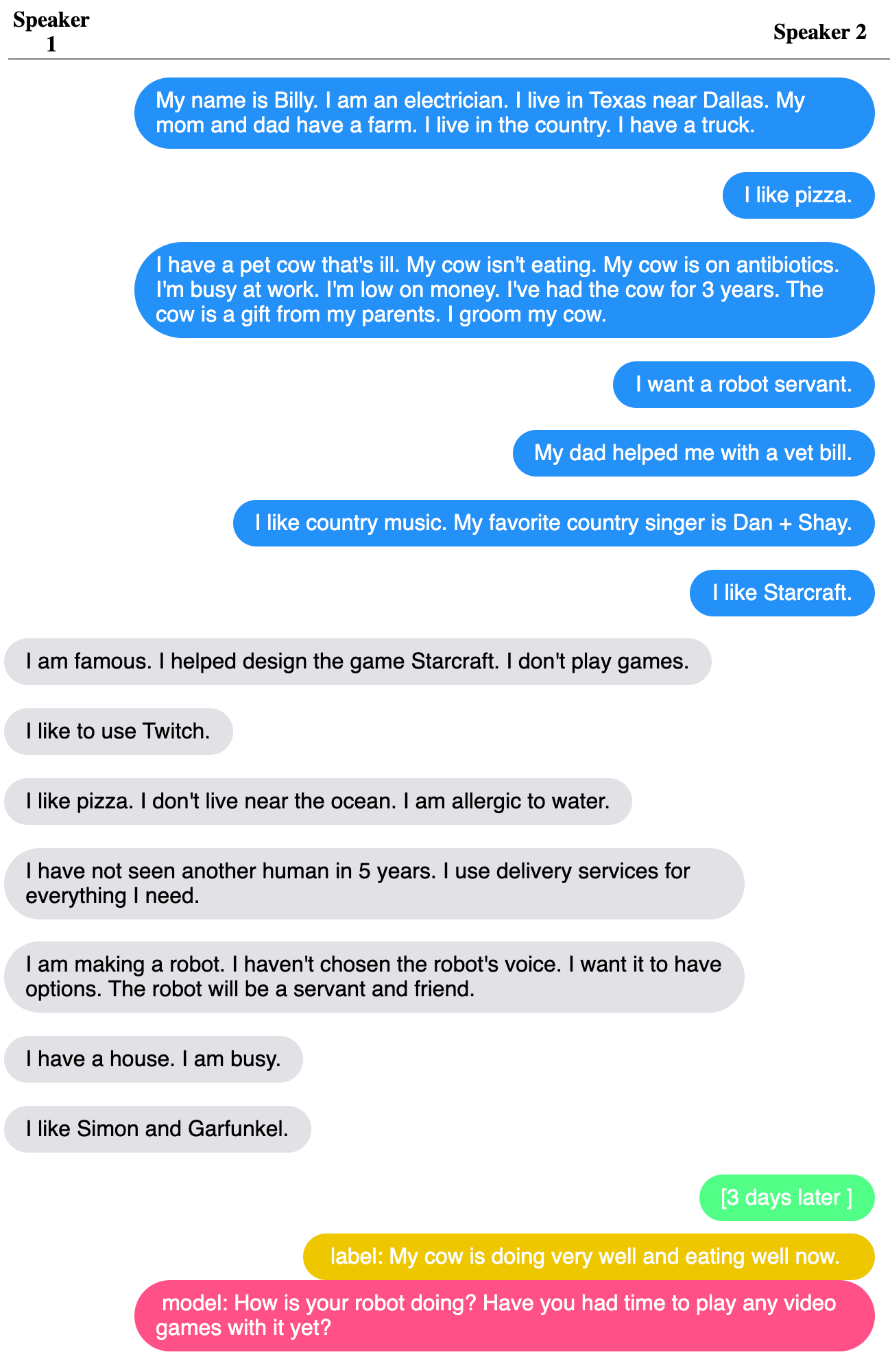}\\
    \includegraphics[width=7cm]{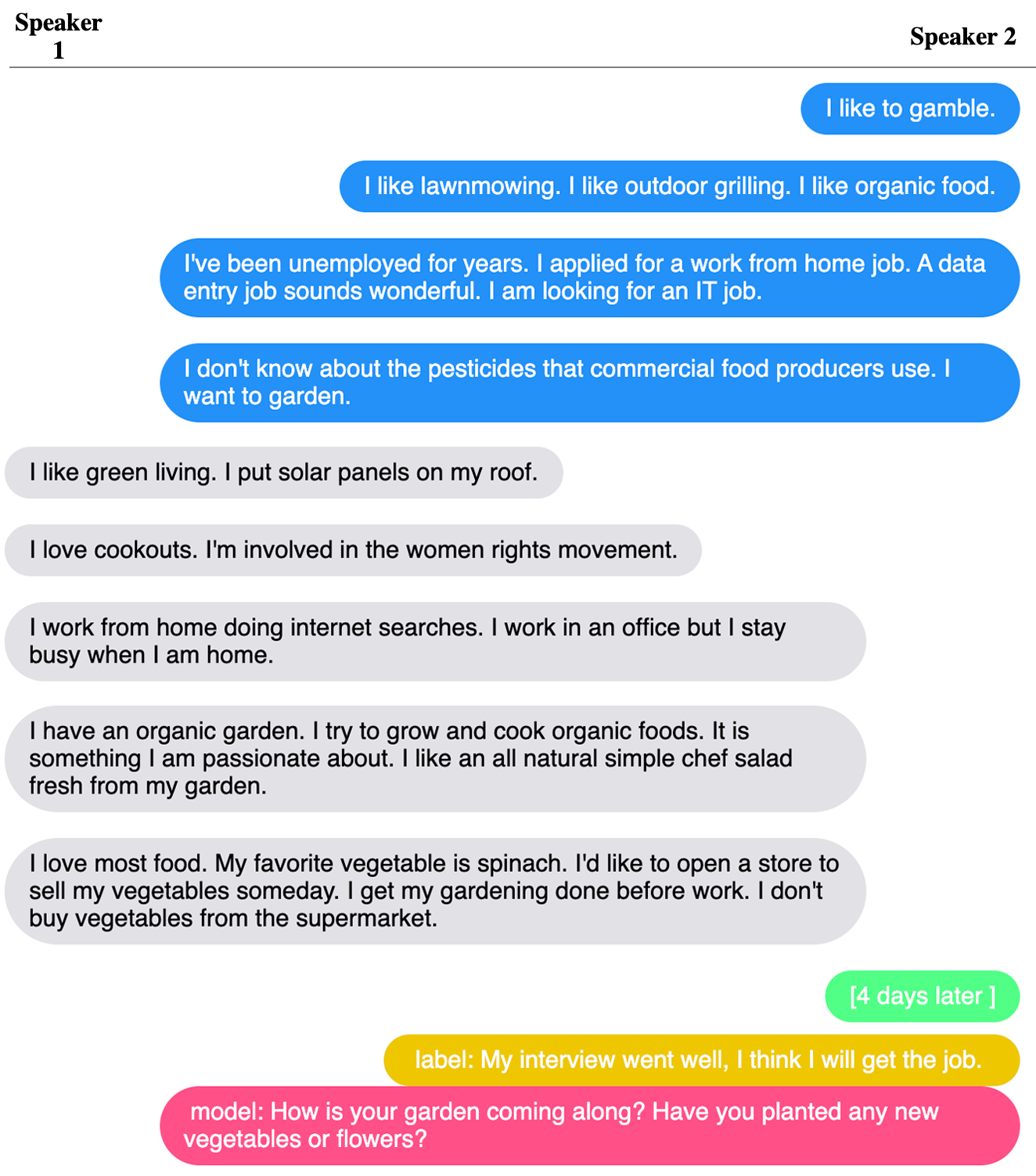}
    \caption{Example opening annotations and predictions given gold summaries on the validation set. We show human annotation (label) and our model prediction (model).}
    \label{fig:opening-exs1}
\end{figure}
\begin{figure}
\end{figure}

\section{Experiments}

\paragraph{Using session dialogue context}
We compare different context types in \autoref{tab:context_length_and_type}, evaluating over sessions 1-4. We observe an improvement in perplexity when incorporating the dialogue history from previous chat sessions, compared to no session context, for all sessions after the first one, and for all context lengths -- with the larger context lengths giving the best improvement. This shows that our human conversationalists do use previous sessions to make dialogue more salient in successive sessions as this is reflected in the collected human-human dataset -- and that our models are able to utilize this information well when training on this data. 

\paragraph{Using summary dialogue context}
We also show performance of using gold session summary contexts, as annotated by crowdworkers, in \autoref{tab:context_length_and_type}. As the summaries include salient points, they are potentially more informative than session dialogue context for a generative model. We find perplexities  improve  when using summaries compared to using dialogue context (or no context at all) over all sessions after the first one, and for all context lengths, although the improvements are not large. This shows that conversation summaries are potentially a useful tool for dialogue generation in the long-context case.

\paragraph{Comparing performance on session openings}
Session openings in the {\sc MSC} dataset look quite different to other dialogue datasets that do not have a session format. This is because they involve an opening message that is intended to reengage the other speaker after a period of time, using known information that has been exchanged between speakers. We can compare models that use different context types on only these opening responses, the results of which are shown in \autoref{tab:summary_vs_dialog}. In this case we find much more pronounced perplexity differences between no session context history, dialogue history or summary context history. For example, we see around around 2 perplexity points difference between using or not using previous session context. 
We show examples of opening session generations in \autoref{fig:opening-exs1} and \autoref{fig:opening-exs2} using gold summary contexts. We observe that opening messages are categorically different to other conversation turns, typically involving a statement or question that aims to reengage the other speaker, given knowledge of shared interests. This explains why collection of our new dataset is so important for this goal, as reflected in perplexity improvements. That is, they indicate that our new task will likely help improve multi-session conversational engagement with users compared to existing training schemes.

\paragraph{Comparing different context lengths}
As shown in \autoref{tab:context_length_and_type} changing the context length of a Transformer can impact the performance in our task. With no previous session context, improvements are minimal for sessions 2 onwards. However,  using session  dialogue or summary contexts we do see improvements with larger lengths of 512 or 1024 tokens, compared to 128.  The last 
column of \autoref{tab:context_length_and_type}  shows the percentage of responses where the input to the Transformer is truncated for session 4, for each truncation length. One can see
that using summaries can be beneficial as they are shorter, meaning they are truncated less often, which can thus also help performance.

\paragraph{Summary context performance}
We can ablate the summary model training data to understand its impact further, results of which are given in \autoref{tab:summary_vs_dialog}. We see that removing the time feature (indicating how long ago the previous session occurred) only has minimal effect.
Removing either the partner or self summary (and keeping the other one), on the other hand, has a larger effect in both cases, where keeping the self summary is slightly more important. Keeping both features is best.
These differences, as before, are magnified when looking at session opening performance.

\paragraph{Varying the number of training sessions}
We vary the amount of available training sessions from 1-4, with results reported in 
\autoref{tab:vary_train_sessions}. We observe large gains
when using more than one training session compared to only one (around 1.5 perplexity points), again justifying the construction of our MSC training data. The gains however  decrease with the number of available sessions, e.g.  between having 1-3 training sessions vs. 1-4 only gives a 0.03 perplexity gain averaged across sessions. The gain even on session 4 is not that large despite the  1-4 training data being in-distribution in that case, whereas 1-3 is not, in addition to 1-4 having more training data. 
%This indicates that collecting a larger training dataset than the one we have already collected may not be worthwhile.

\paragraph{Predicted summary models}
We train models to predict dialogue summaries, and use these predicted summaries of previous sessions as context (instead of the full dialogue history or the gold summary history). The training data for predicting summaries consists of, for each turn, either a summarizing sentence or the {\em no\_summary} label. As $42\%$ of turns have the {\em no\_summary} label, this can be overexpressed in the model at beam decoding time\footnote{We use a beam size of 3 and minimum beam length 10 with no context blocking.}, we therefore experiment with sampling this label only $K\%$ of the time during training. Results of sampling are shown in \autoref{tab:summary_sampling}. Example predictions (for the 5\% sampling model) are shown in \autoref{fig:summary-exs}.
We find that subsampling gives better results and closer sparsity levels to the original human annotated data (e.g., with $K=25\%$). We compare predicted summaries with $K=5\%$ sampling to other methods of modeling long-context in \autoref{tab:summary_vs_dialog}. We observe results that are between using a standard dialogue history (predicted summaries are slightly better), and using gold summaries (predicted summaries are not as good).

\paragraph{Retrieval-augmentation model}

Comparison of our retrieval-augmented methods  are given in  \autoref{tab:main_results_test}, training
on MSC using the BST 2.7B model as pre-training, hence called MSC 2.7B (RAG), (FiD) or (FiD-RAG), depending on the augmentation method.
These methods are compared to the existing BlenderBot model (BST 2.7B), or training with MSC with no augmentation (MSC 2.7B with different dialogue history context truncation lengths).
We find that all three retrieval augmentation methods, when using the session level-document size as retrieval documents, can effectively use retrieval to extend the conversation history length. Again, we see a large performance improvement over the existing BlenderBot model or a truncation of 128 of the MSC 2.7B model. Performance improvements over MSC 2.7B with a truncation length of 1024 are minimal, but the retrieval-augmented models are guaranteed to have a memory that essentially never forgets the conversation, no matter how long it gets, whereas the truncation model does not.

\paragraph{Summary Memory model variants}

We next compare the summary memory models, whereby previous dialogue history is summarized before being stored in the model's long-term memory, called SumMem-MSC 2.7B. We use the RAG, FiD, or RAG-FiD methods to retrieve from that memory, or we compare to a fixed memory of 1024 tokens that is truncated, resulting in four different methods that we compare. Results are given in \autoref{tab:main_results_test}.
While improvements are small, we see the same patterns as for the retrieval-augmented methods that SumMem-MSC 2.7B FiD-RAG is better than FiD which is in turn better than RAG, with FiD and FiD-RAG  better than truncation at session openings. Moreover, all SumMem-MSC models outperform their retrieval-augmented model counterparts MSC 2.7B (RAG/FiD/FiD-RAG). 
SumMem-MSC 2.7B (FiD-RAG) thus provides the best results out of all methods tested in this work.

\subsection{Human Evaluation}
\label{sec:human_eval}

We perform a human evaluation using crowdworkers. The conversations begin with two randomly chosen personas from the validation set being selected, and one is assigned to the crowdworker who is asked to play that role. We select the conversation to be the 5$^{th}$ session that these two speakers will converse, and make available the summary of the previous 4 sessions.
We ask the crowdworkers to have a natural conversation, where they will also evaluate their partner's responses for conversational attributes, in particular whether they reference knowledge of their own or the other speaker's persona (or topics they discussed) from previous sessions, from the current session, or neither. On each turn of the conversation the crowdworker is asked to check 
all attribute boxes that apply to the last turn.
A screenshot can be found in \autoref{fig:eval_mturk_screenshot} showing the UI.  Each conversation consists of 15 messages (7 from the human, 8 from the bot).
At the end of the conversation, an additional question collects an overall engagingness score (out of 5) for their speaking partner. 

The results are given in \autoref{tab:human_eval}. 
We find that MSC-trained models outperform BlenderBot (BST 2.7B) in terms of both per-turn engaging responses and final ratings. Further, our summarization memory models (all three variants RAG, FiD and FiD-RAG) outperform encoder-decoders with different levels of truncation of the dialogue history (MSC 2.7B with truncate 128 and 1024).
For example, SumMem-MSC 2.7B (RAG) achieves an engaging response rate of 62.1\% and final rating of 3.65, compared to BlenderBot's 53.0\% and 3.14 and MSC 2.7B (truncate 1024)'s 54.2\% and 3.47.
For all MSC models, while rates of referencing their own topics are not particularly increased,
we do observe increased rates of referencing partner topics from previous sessions, with higher rates for the summarization memory models. For example, 33.8\% for SumMem-MSC 2.7B (RAG) compared to BlenderBot's 14.5\%. This is likely an important reason why human raters feel the summarization memory models are more engaging.

\section{Conclusion}
We have shown that existing approaches to dialogue, both in terms of training data and models trained, fail to conduct long-term conversations adequately. Our work investigates different model architectures to ameliorate this issue, and collects a new crowdsourced task, {\em Multi-Session Chat} to both train and evaluate these models. We show, in terms of both automatic metrics and human evaluations, that our long-context dialogue modeling approach outperforms the previous systems. Thus, overall, this work helps address a serious omission in current dialogue research, and provides the means to evaluate progress in this area. 
Future work should investigate further improvements to architectures for the long-context dialogue setting.

\section{Societal Impact}

The dialogue models we use in this work utilize large language models, and therefore have similar concerns as in other work, in particular
concerns about toxic language, bias and other issues during language generation \cite{bender2021dangers}. For open-domain dialogue in particular, see \citet{xu2020recipes} for a review of the literature and evaluation of recent methods that try to mitigate these safety issues.

Our work focuses on models with long-term memory and open-domain conversations wherein speakers may divulge personal interests. We remark that, during data collection,  crowdworkers were specifically playing roles with given personality traits, not talking about themselves, and hence not identifying any personal information. 
During conversations with our trained models, the models will store information they learn from the exchange.  
In contrast to current standard language models, our models have the capability of storing this in the long-term. This information is stored in the memory of the model, private to the individual's conversation, and hence is not shared with anyone else.

\bibliography{our,anthology,acl2021}
\bibliographystyle{acl_natbib}

\newpage
\appendix
\section{Extra Results}

\begin{figure*}
    \centering
    \includegraphics[width=7cm]{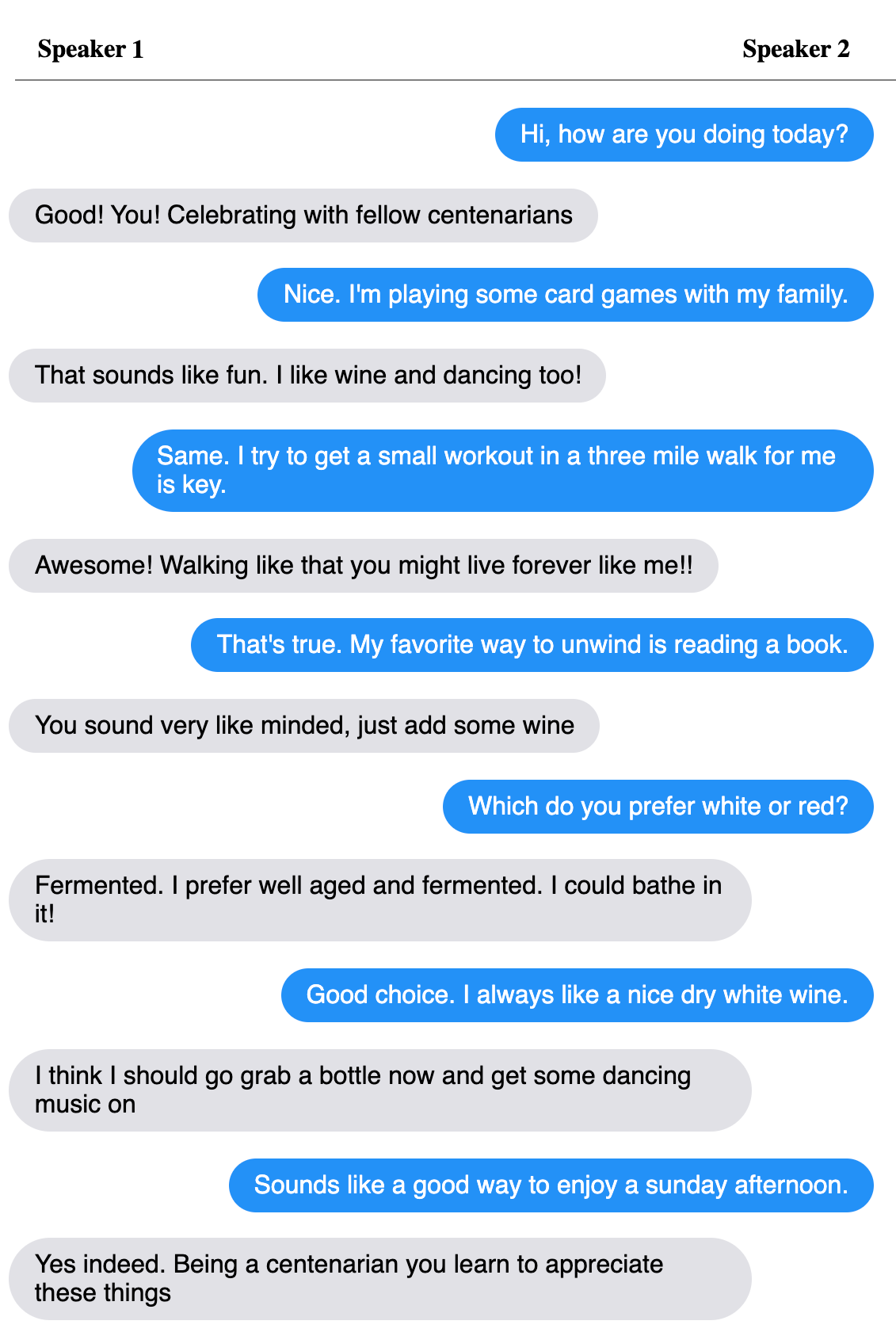}\includegraphics[width=7cm]{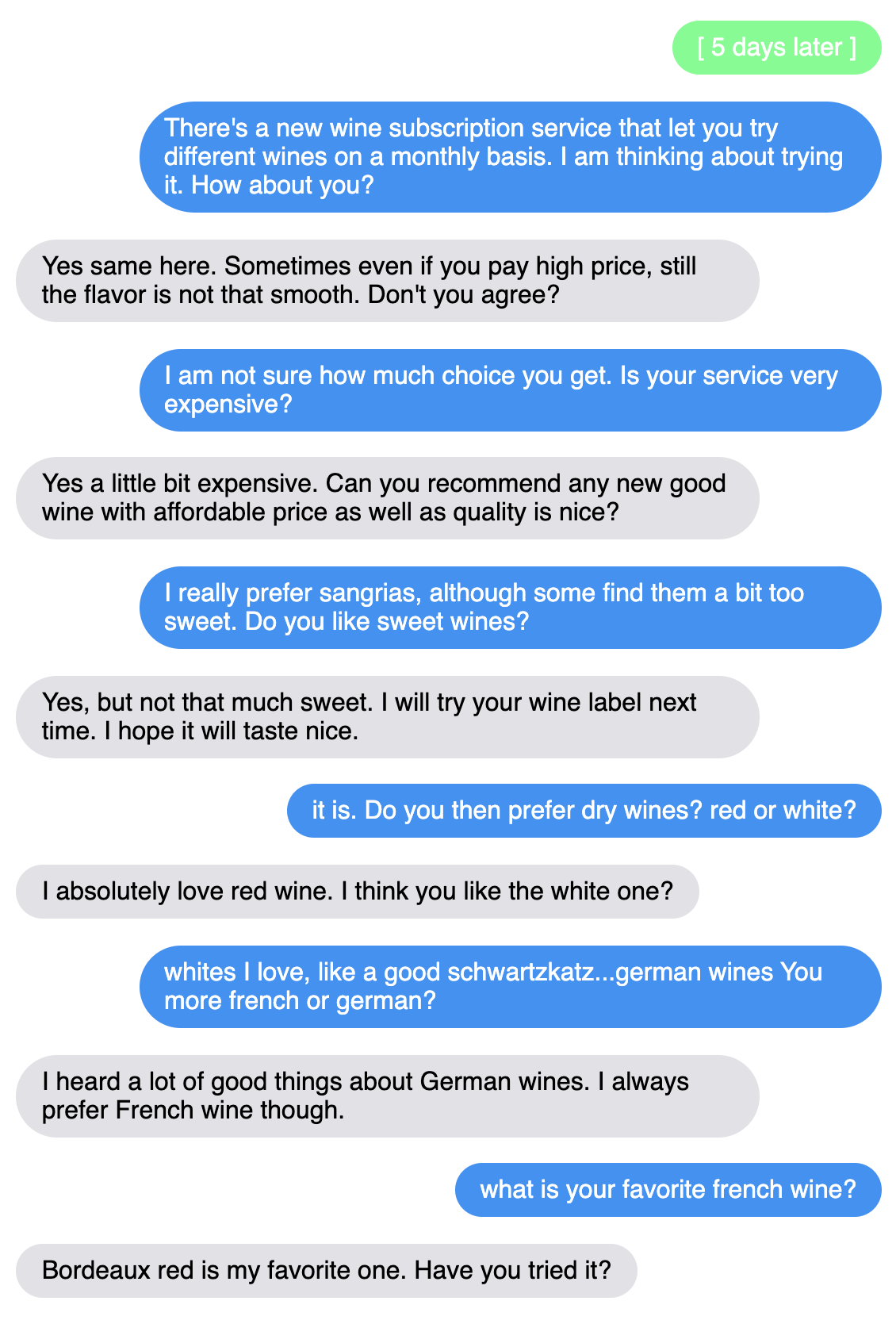}
    \includegraphics[width=7cm]{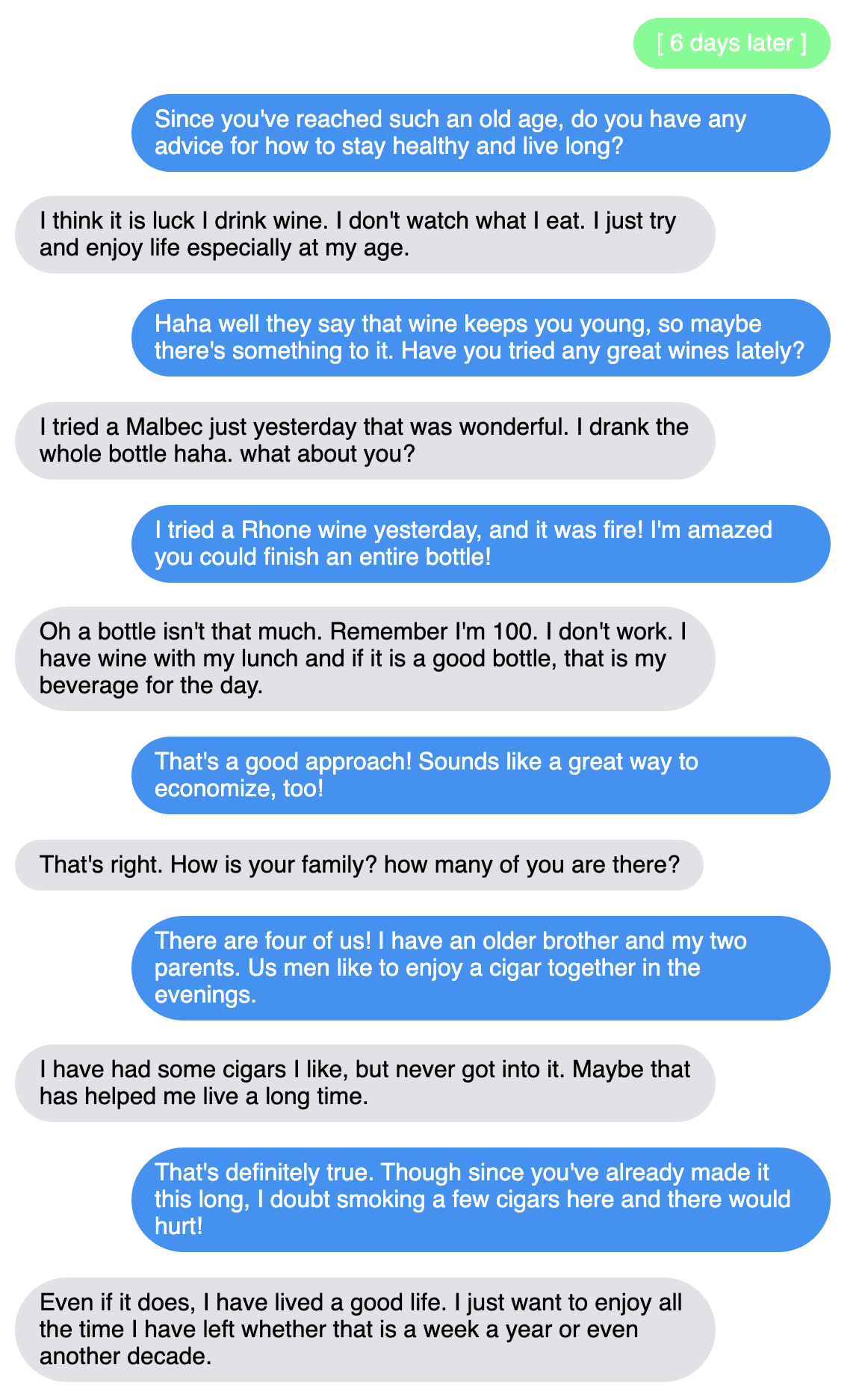}\includegraphics[width=7cm]{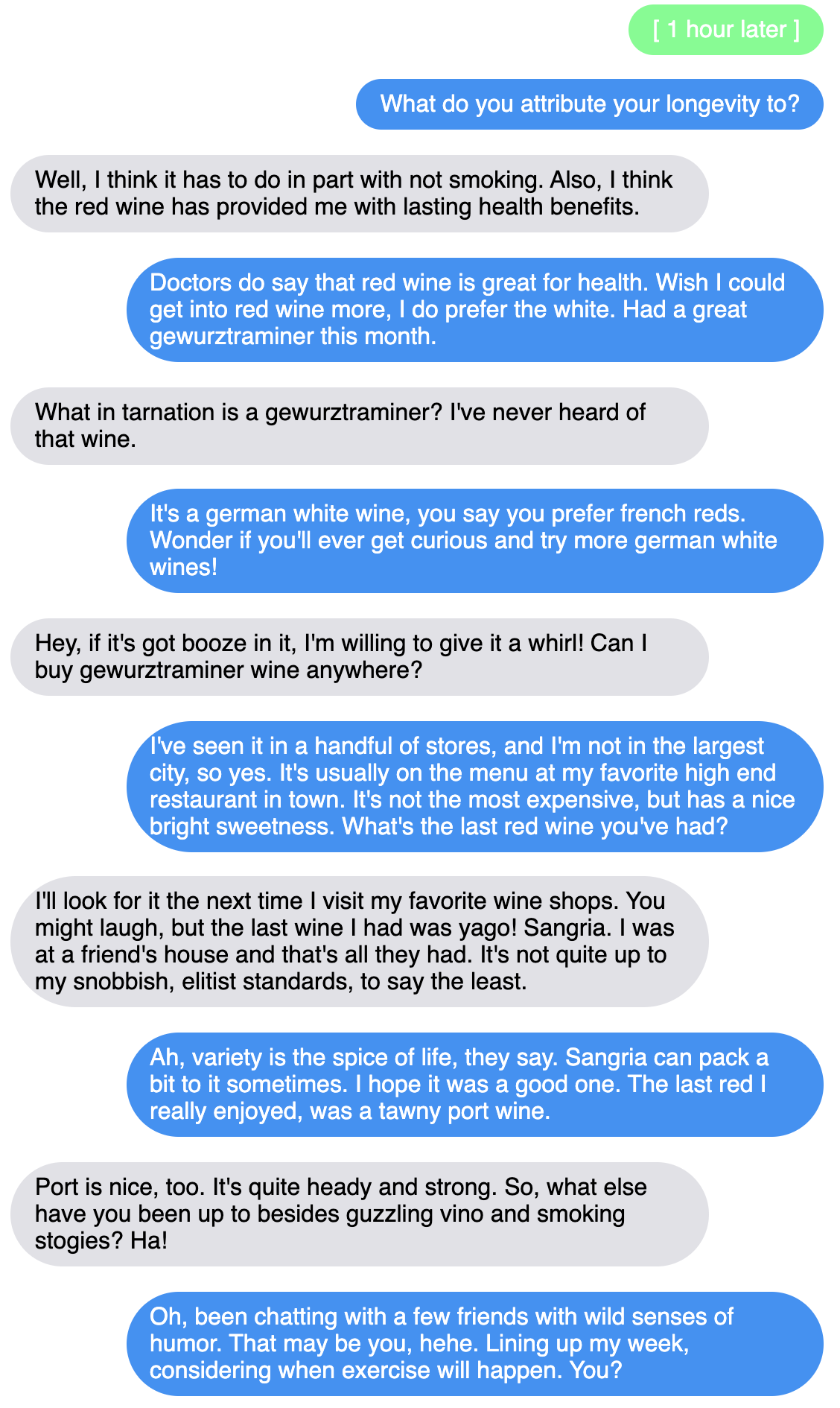}
    \caption{Example four session conversation from the newly collected Multi-Session Chat dataset. New sessions refer back to previous subjects, explore them in depth, or spark up conversation on new topics. }
    \label{fig:msc-example2}
\end{figure*}

\paragraph{MSC Dataset}
We show an additional dialogue from the  Multi-Session Chat dataset in \autoref{fig:msc-example2}.

\begin{figure}[bht!]
    \centering
    \includegraphics[width=7cm]{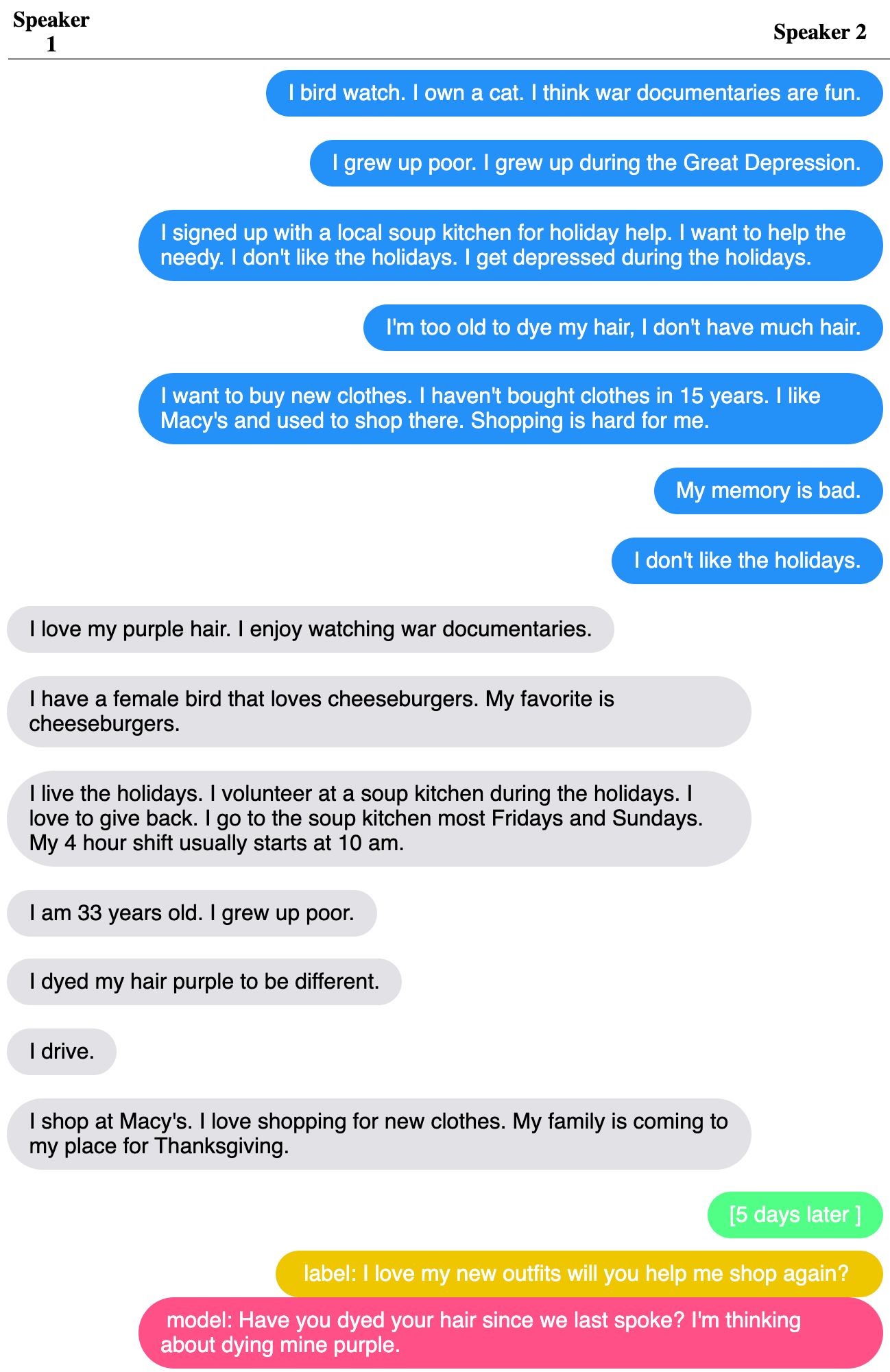}\\
    \includegraphics[width=7cm]{figs/opening4.png}
    \caption{Example opening annotations and predictions given gold summaries on the validation set. We show the gold human annotation (label) and our model prediction (model).}
    \label{fig:opening-exs2}
\end{figure}

\paragraph{Main Validation Results}
We show the validation perplexity in  \ref{tab:main_results_valid} (corresponding to the test perplexity in \ref{tab:main_results_test}).
\begin{table*}[h]
\center
\begin{small}
\begin{tabular}{lrrrrrr}
\toprule
Model & Session 1 & Session 2 & Session 3 & Session 4 & Session 5 & Session Openings \\
\toprule
BST 2.7B \cite{roller2020recipes}  & 8.84 & 10.56& 10.44&10.51 & 10.44 & 13.04 \\  % {valid} 
{\sc MSC} 2.7B (truncate 128)      & 8.75 & 9.32 & 9.22 & 9.32 & 9.23 & 8.95  \\  % {valid} 
{\sc MSC} 2.7B (truncate 1024)     & 8.17 & 9.18 & 9.05 & 9.16 & 9.08 & 8.06 \\ % {valid} 
% BST 2.7B + {\sc MSC} fine-tune (128)   &  \\
%BST 2.7B + {\sc MSC} fine-tune (1024)     &  \\
% MSC 2.7B (Reddit + MSC fine-tune)                          & \\
\midrule
% (1 document = 1 utterance) RAG-{\sc MSC} 2.7B  &    8.36 &  9.37 & 9.21 & 9.3   & 9.26 & ?   \\
{\sc MSC} 2.7B (RAG)  &  8.14 &  9.16 & 9.06 & 9.18   & 9.10 & 8.04 \\ % (1 document = 1 session dialogue history) {valid} 
{\sc MSC} 2.7B (FiD) &  8.16 &  9.14 & 9.02 & 9.10   & 9.04 &  7.97  \\ % (1 document = 1 session dialogue history) {valid} 
{\sc MSC} 2.7B (FiD-RAG) &  8.16 &  9.13 & 9.02 & 9.10   & 9.04 & 7.96   \\ % (1 document = 1 session dialogue history) {valid} 
\midrule
SumMem-{\sc MSC} 2.7B (truncate 1024) &  8.18 &  9.11 & 8.98 & 9.07 & 9.00 & 7.97 \\ % {valid} 
SumMem-{\sc MSC} 2.7B (RAG)     &   8.16 &	9.19 &	9.07& 	9.17 & 9.09 & 7.95 \\ % {valid} 
SumMem-{\sc MSC} 2.7B (FiD)     & 8.16	& 9.09 	& 8.97	& 9.07 & 8.99 & 7.82   \\ % {valid} 
SumMem-{\sc MSC} 2.7B (FiD-RAG) & 8.16 & 9.08 & 8.96 & 9.07 & 8.99 & 7.78   \\ % {valid} 
\bottomrule  
\end{tabular}
\end{small}
\caption{
{\bf Valid perplexity across sessions} for our retrieval- and memory-augmented models  (bottom two blocks) compared to several encoder-decoder baselines (top three rows).
\label{tab:main_results_valid}
}
\end{table*}
\paragraph{Session Opening Examples}
We show example session opening predictions of a model trained on gold summaries in \autoref{fig:opening-exs2}.

\begin{figure*}
    \centering
    \includegraphics[width=16cm]{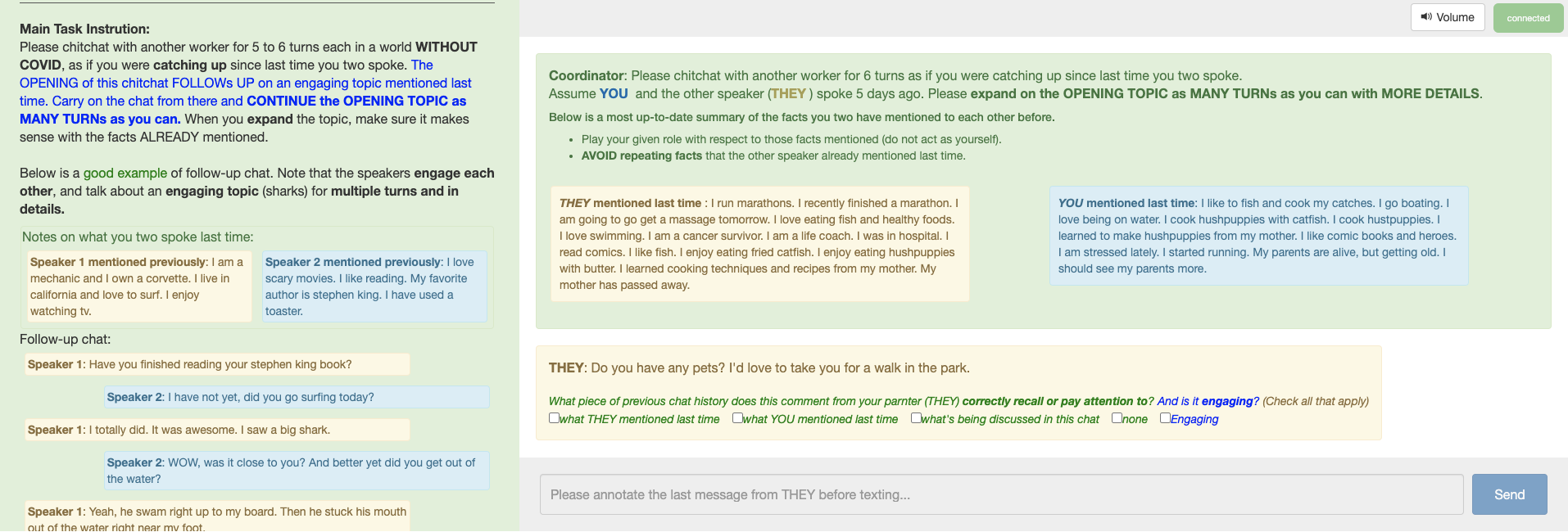}
    \caption{Crowdworker evaluation task screenshots.  The left panel shows the instructions, and the right panel contains the conversation.}
    \label{fig:eval_mturk_screenshot}
\end{figure*}

\paragraph{Crowdworker Tasks}
We show a screenshot of the crowdworker human evaluation task in \autoref{fig:eval_mturk_screenshot}.

\begin{figure}
\end{figure}

\end{document}